\documentclass[10pt,twocolumn,letterpaper]{article}

\usepackage[pagenumbers]{iccv} 

\newcommand{\ours}{$\mathbf{S^4M}$\xspace}
\newcommand{\ourkd}{SD\xspace}
\newcommand{\ouraug}{ARP\xspace}
\definecolor{lightgray}{rgb}{0.85,0.85,0.85}
\definecolor{darkgray}{rgb}{0.5,0.5,0.5}
\newcommand{\paragrapht}[1]{\vspace{-10pt}\paragraph{#1}}
\newcommand{\grayx}{\textcolor{lightgray}{\ding{55}}}
\newcommand{\blackcheck}{\ding{51}}

\definecolor{iccvblue}{rgb}{0.21,0.49,0.74}
\usepackage[pagebackref,breaklinks,colorlinks,allcolors=iccvblue]{hyperref}

\usepackage{colortbl}
\usepackage{xcolor}
\usepackage{multirow}
\usepackage{subcaption}
\usepackage{amssymb}
\usepackage{pifont}
\usepackage{blindtext}
\usepackage{array}
\usepackage{cellspace}
\usepackage[utf8]{inputenc}
\usepackage{algorithm}
\usepackage{algpseudocode}
\usepackage{amsmath}
\usepackage{amssymb}
\usepackage{booktabs} 
\usepackage{xcolor} 
\usepackage{colortbl} 
\usepackage{array} 
\usepackage{lipsum}
\usepackage{pifont}

\def\paperID{7890} 
\def\confName{ICCV}
\def\confYear{2025}

\title{$\mathbf{S^4M}$: Boosting Semi-Supervised Instance Segmentation with SAM}

\author{
Heeji Yoon$^{1*}$ \qquad Heeseong Shin$^{1*}$ \qquad Eunbeen Hong$^{2}$ \qquad
Hyunwook Choi$^{2}$ \\ Hansang Cho$^{3}$ \qquad Daun Jeong$^{3}$ \qquad Seungryong Kim$^{1\dagger}$ \\[5pt]
$^{1}$KAIST AI \qquad $^{2}$Korea University \qquad $^{3}$Samsung Electro-Mechanics\\[5pt]
{\tt \href{https:/cvlab-kaist.github.io/S4M}{https://cvlab-kaist.github.io/S4M}}
}

\begin{document}
\maketitle

\def\thefootnote{*}\footnotetext{These authors contributed equally.}\def\thefootnote{\arabic{footnote}}
\def\thefootnote{$\dagger$}\footnotetext{Corresponding author.}\def\thefootnote{\arabic{footnote}}

\begin{abstract}
Semi-supervised instance segmentation poses challenges due to limited labeled data, causing difficulties in accurately localizing distinct object instances. Current teacher-student frameworks still suffer from performance constraints due to unreliable pseudo-label quality stemming from limited labeled data. While the Segment Anything Model (SAM) offers robust segmentation capabilities at various granularities, directly applying SAM to this task introduces challenges such as class-agnostic predictions and potential over-segmentation. To address these complexities, we carefully integrate SAM into the semi-supervised instance segmentation framework, developing a novel distillation method that effectively captures the precise localization capabilities of SAM without compromising semantic recognition. Furthermore, we incorporate pseudo-label refinement as well as a specialized data augmentation with the refined pseudo-labels, resulting in superior performance. We establish state-of-the-art performance, and provide comprehensive experiments and ablation studies to validate the effectiveness of our proposed approach.
\end{abstract}
\section{Introduction}

Instance segmentation—simultaneously detecting objects and delineating their pixel-level boundaries—is fundamental to applications such as autonomous driving and medical imaging~\cite{CERON2022102569,Zhang_2016_CVPR}. Although fully-supervised methods~\cite{he2017mask, cheng2021per, cheng2022masked, hong2022cost} have achieved impressive accuracy, their dependence on extensive annotated datasets limits scalability due to the labor-intensive nature of pixel-level labeling. 

Consequently, recent work~\cite{wang2022noisy, Berrada_2024_WACV, chen2024depth, 10460559, hu2023pseudo} has explored semi-supervised learning (SSL) approaches that additionally leverage unlabeled images. Typically, these methods generate pseudo-labels using a teacher network trained on a limited labeled dataset, which subsequently guides the training of a student network. However, the restricted availability of labeled data renders the pseudo-labels error-prone, thereby reducing the benefits of incorporating unlabeled images in the teacher-student framework. 

To address this, recent studies have proposed several strategies, including noise filtering~\cite{wang2022noisy}, the use of auxiliary information (e.g., depth maps~\cite{chen2024depth}), and dedicated training stages to stabilize the teacher–student framework~\cite{10460559, Berrada_2024_WACV}. Despite these advances, the inherent limitations imposed by the small-scale labeled data continue to produce pseudo-labels with significant errors, ultimately impeding the performance of the student network.

\begin{figure}[t]
    \centering
    \includegraphics[width=\columnwidth]{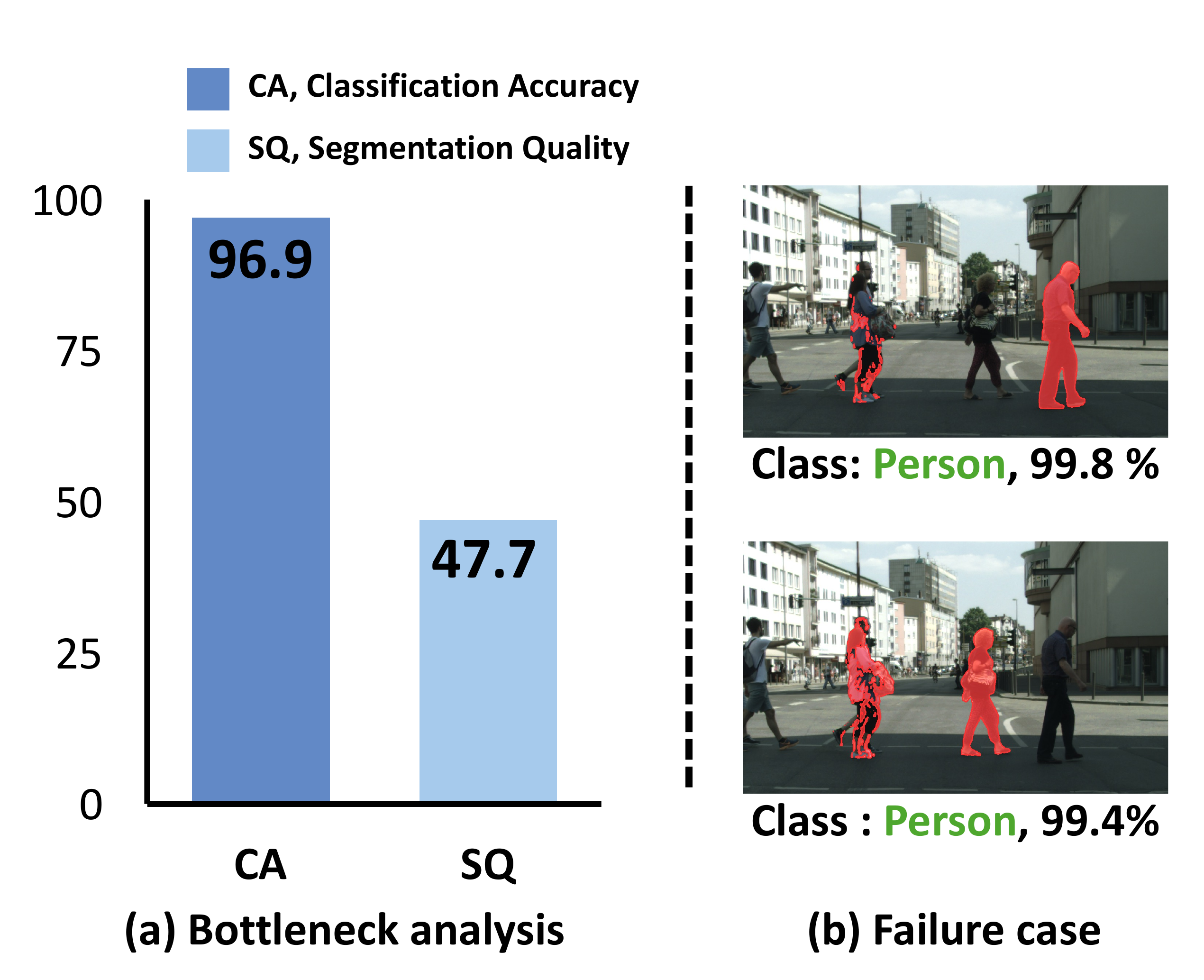}
    \vspace{-15pt}
    \caption{\textbf{Analysis on  pseudo-labels by the teacher in a teacher-student framework for semi-supervised instance segmentation.}  (a) Bottleneck analysis revealing that the primary limitation lies in mask quality rather than classification. Note that class accuracy (CA) is computed on matched pairs with IoU $>$ 0.5, and segmentation quality (SQ) is measured by the standard segmentation quality metric from panoptic quality~\cite{kirillov2019panoptic}. (b) Example failure cases with correct, confident class prediction but inaccurate masks.}
    \label{fig:motivation}
    \vspace{-10pt}
\end{figure}

Recently, large-scale vision foundation models~\cite{oquab2023dinov2, he2022masked, radford2021learning, weinzaepfel2022croco}, pretrained on web-scale datasets, have demonstrated exceptional performance across diverse tasks and exhibit strong generalization without task-specific fine-tuning~\cite{amir2021deep, an2024cross, zhou2022extract}. In particular, the Segment Anything Model (SAM)~\cite{kirillov2023segment, ravi2024sam2} has gathered significant attention as a prompt-driven segmentation foundation model, capable of predicting fine-grained masks at any granularity, from whole objects to parts and sub-parts, using geometric prompts such as points and bounding boxes. Through training on an unprecedented scale of images, SAM has demonstrated its efficacy and generalization capabilities in various domains~\cite{huang2024segment, ren2024grounded, cao2023segment}, as well as its application in diverse tasks~\cite{yang2023track, rajivc2023segment}. 

These advancements motivate our exploration of SAM to enhance instance segmentation through knowledge distillation, pseudo-label enhancement, and data augmentation, which are fundamental elements in semi-supervised learning. However, it still faces challenges in directly applying SAM to instance segmentation due to its class-agnostic design~\cite{kirillov2023segment, wang2024sam}: instance segmentation inherently requires both mask and class predictions, yet SAM is not designed to generate class-conditioned masks. 

In this work, we integrate SAM into the semi-supervised instance segmentation framework to address challenges associated with limited labeled data. We first examine the deficiencies of existing semi-supervised approaches~\cite{Berrada_2024_WACV} by visualizing pseudo-labels generated by their teacher networks, thereby establishing a basis for incorporating SAM. As illustrated in Fig.~\ref{fig:motivation}, while these networks reliably identify classes, they often fail in precise localization by grouping multiple instances into a single mask, which we refer to as \textit{under}-segmentation. Although it might appear straightforward to use SAM to separate instances within these pseudo-labels, its class-agnostic design frequently produces masks that capture only fine-grained segments of an object rather than the object as a whole, resulting in \textit{over}-segmentation~\cite{shin2025towards}. This underscores the necessity for carefully balancing under- and over-segmentation when integrating SAM into the semi-supervised instance segmentation framework.

In this regard, we argue that it is crucial to identify what and what not to learn from SAM for tackling the under- and over-segmentation problem, and propose a novel framework for distilling SAM to improve the teacher and student networks. In specific, we first improve the teacher network trained on a small amount of label data by a novel knowledge distillation objective, which can effectively acquire the fine-grained localization capabilities of SAM while avoiding over-segmentation or hindering semantic recognition. We further enhance our framework by propagating the strong segmentation capability of SAM through pseudo-label refinement and an augmentation strategy designed for instance segmentation. 

\begin{figure*}[ht]
    \centering
    \includegraphics[width=1.0\linewidth]{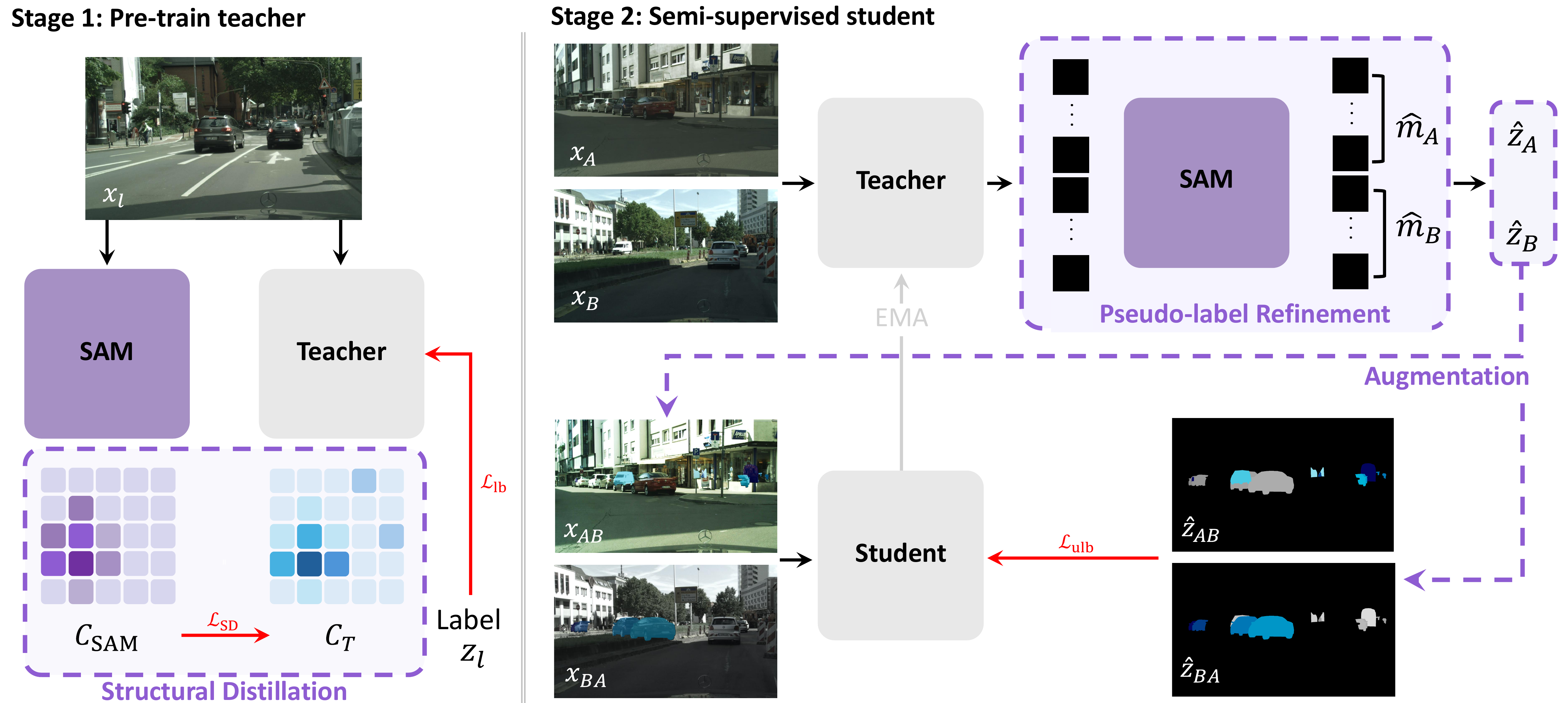}
    \caption{\textbf{Overall pipeline of the proposed framework, \ours.} We propose \ours, a semi-supervised instance segmentation framework that effectively leverages SAM knowledge through three key approaches. First, we improve the teacher network through structural distillation, which distills SAM's inherent spatial understanding. Then, as the student learns from unlabeled images, we apply pseudo-label refinement based on SAM's strong segmentation capability, and further enhance training with instance-aware augmentation, \ouraug, which leverages the improved pseudo-labels.}
    \label{fig:main_figure}
    \vspace{-5pt}
\end{figure*}

Our framework for boosting \underline{\textbf{S}}emi-\underline{\textbf{S}}upervised Instance \underline{\textbf{S}}egmentation with \underline{\textbf{S}}AM, or \ours, establishes state-of-the-art performance in all benchmarks, demonstrating the effectiveness of our approach. We further provide detailed ablations and analysis on our methodology, as well as qualitative comparison with baselines.

Our main contributions can be summarized as follows:
\begin{itemize}
    \item We incorporate Segment Anything Model (SAM) into the semi-supervised instance segmentation framework, and present our explorations for improving the teacher-student framework with SAM.
    \item We carefully design our framework to fully leverage SAM through structural distillation, pseudo-label refinement and data augmentation while avoiding potential drawbacks of directly adopting SAM.
    \item We establish state-of-the-art performance across benchmarks, and provide thorough ablations and analysis to validate our approach.
\end{itemize}

\section{Related Work}

\paragraph{Semi-supervised instance segmentation.}
Dominant approaches to semi-supervised instance segmentation~\cite{Wang_2022_CVPR,10460559,hu2023pseudo,Berrada_2024_WACV,chen2024depth} have been based on student-teacher pseudo-labeling, where teacher model generates pseudo-labels for unlabeled images, which are then used by student model for training. In this framework, applying weak and strong data augmentations for generating and predicting pseudo-labels, respectively, effectively utilizes unlabeled data, as proposed in FixMatch~\cite{sohn2020fixmatch}. The semi-supervised instance segmentation task was first introduced by Noisy Boundaries~\cite{Wang_2022_CVPR}, which proposed a noise-tolerant mask head to filter noisy student predictions. Polite teacher~\cite{10460559} employs EMA teacher while filtering out pseudo-labels by confidence thresholding. PAIS~\cite{hu2023pseudo} introduced a dynamically changing loss weight based on the quality of pseudo-labels, improving the utilization of unlabeled data by retaining low-confidence labels. 

More recently, GuidedDistillation~\cite{Berrada_2024_WACV} proposed a guided burn-in stage, to improve the distillation approach. The method first trains the teacher model on labeled data and then independently trains the student model with both labeled and unlabeled data before the main training. However, previous methods still struggle to generate noise-tolerant masks due to the limited amount of labeled data. To address this, Depth-Guided~\cite{chen2024depth} incorporates a depth foundation model into the student-teacher framework for improved understanding. In this work, we integrate SAM into a semi-supervised instance segmentation framework for the first time, fully leveraging its powerful generalization capabilities.

\paragrapht{Segment Anything Model.}
SAM~\cite{kirillov2023segment,ravi2024sam2} is a foundation model for image segmentation, designed to perform zero-shot segmentation across diverse domains without task-specific fine-tuning. As an interactive segmentation model, SAM takes an image along with a set of prompts, such as points, bounding boxes, masks, or a combination of these, as input, enabling flexible and instance-aware segmentation. This flexibility allows SAM to generalize effectively across a wide range of segmentation tasks. While SAM excels at easily segmenting objects or regions, it has limitations in understanding objects in the broader context of a scene. We leverage strong delineation capability of SAM to improve the separation of instances, particularly in situations where the labels may have a more semantic focus.

\paragrapht{Knowledge distillation.}
Knowledge distillation (KD) is a widely known technique for transferring knowledge from a teacher to a student model~\cite{hinton2015distillingknowledgeneuralnetwork}. \cite{Zagoruyko2016PayingMA} proposed transferring attention maps to guide the student model in mimicking spatial focus of the teacher, emphasizing the importance of a more spatially aware distillation approach. In the domain of segmentation, several studies have investigated KD methods for semantic segmentation~\cite{10858405,Liu_2019_CVPR,He2019KnowledgeAF}, with an emphasis on distilling spatial relationships.

However, extending spatial distillation to instance segmentation has been less explored, from its added complexity of distinguishing different instances as well as classification. Capturing structured information is especially critical in instance segmentation, where distinguishing instances in closely related regions requires detailed structural understanding. Our structural distillation approach is specifically designed to leverage the structural cues from SAM, while avoiding the transfer of undesirable traits—such as limited semantic understanding~\cite{wang2024sam, vs2024possam}—that could undermine instance segmentation. This motivates our exploration of what to distill from SAM and what to omit, in contrast to previous methods in semantic segmentation~\cite{10858405,Liu_2019_CVPR,He2019KnowledgeAF}.

\section{Preliminaries}

\paragraph{Problem formulation.}
\label{Formulation}
In semi-supervised instance segmentation, we leverage a large set of unlabeled data $\mathcal{D}_U = \{x_u\}$ and a small labeled data set $\mathcal{D}_L = \{(x_l, z_l)\}$, where each image $x\in \mathbb{R}^{3\times H\times W}$ has a spatial resolution of height $H$ and width $W$. The ground truth $z_l=\{(c^k_l, m^k_l)\}$ for labeled image $x_l$ consists of class labels $c^k \in \{1, ..., K\}$ and binary masks $m^k \in \{0, 1\}^{H \times W}$, where $k$ indexes each instance in $x_l$.
The goal is to improve model performance beyond what $\mathcal{D}_L$ alone provides. 

Our \ours is built on the widely adopted teacher-student framework with consistency regularization~\cite{Berrada_2024_WACV, chen2024depth} that formulates the training pipeline into two stages~\cite{Berrada_2024_WACV, chen2024depth}. In the first stage, we pre-train the teacher $\mathcal{F}_T$ on labeled data $\mathcal{D}_L$ with the objective $\mathcal{L}_T = \mathcal{L}_\mathrm{lb}$. After obtaining the pre-trained teacher network, we then train the student $\mathcal{F}_S$ utilizing both labeled data $\mathcal{D}_L$ and unlabeled data $\mathcal{D}_U$. At the second stage, the teacher $\mathcal{F}_T$ processes weakly augmented views $\text{weak}(x_u)$ to generate pseudo-labels $\hat{z}_u=\{(\hat{c}_u^k, \hat{m}_u^k)\}$, retaining predictions where class confidence exceeds $\tau_c$. The student $\mathcal{F}_S$ then learns from strongly augmented views $\text{strong}(x_u)$ by matching its predictions to $\hat{z}_u$. 

The total objective is defined as $\mathcal{L}_S = \mathcal{L}_\mathrm{lb} + \lambda_\mathrm{ulb}\mathcal{L}_\mathrm{ulb}$. The network is jointly optimized by a cross entropy loss $l_\mathrm{cls}$ for classification and a mask loss $l_\mathrm{mask}$ consisted of dice loss~\cite{milletari2016v} and binary cross entropy. Consequently, we define $
\mathcal{L}_\mathrm{lb}=l_\mathrm{cls}(\tilde{c}_l^k, c^k_l)+ \lambda_\mathrm{mask}l_\mathrm{mask}(\tilde{m}_l^k, m^k_l)$ and $\mathcal{L}_\mathrm{ulb} = l_\mathrm{cls}( \tilde{c}^k_u, \hat{c}^k_u)+ \lambda_\mathrm{mask}l_\mathrm{mask}(\tilde{m}^k_u, \hat{m}^k_u)$, where $\{(\tilde{c},\tilde{m})\}$ are model predictions.

\paragrapht{Baseline segmentation network.} Our framework builds on Mask2Former~\cite{cheng2022masked}, a unified architecture for segmentation that we adapt for instance segmentation, following Guided Distillation~\cite{Berrada_2024_WACV}. The model comprises three core components. An image encoder extracts low-resolution features from the input image, and a pixel decoder progressively upsamples and refines these features to construct a multi-scale feature pyramid. A transformer decoder processes $N$ learnable query embeddings, where  iteratively interacting with with the multi-scale features to generate class embeddings for each of the $N$ segments. The binary masks for each segment are generated by computing the dot product between the segment embeddings and the per-pixel embeddings, followed by a sigmoid activation. 

\paragrapht{SAM.}
The SAM~\cite{wang2024sam, ravi2024sam2} is architecturally structured around three core components: an image encoder, a prompt encoder, and a mask decoder. Image encoder utilizes ViT-based backbone~\cite{dosovitskiy2020image,ryali2023hiera} to extract image features and generate $H'\times W'$ spatial embedding, where $H'$ and $W'$ denote the height and width of the feature map, respectively. The prompt encoder captures interactive positional cues from inputs such as points, boxes, and masks, transforming them into prompt embedding that inform the segmentation process. The final mask decoder fuses image and prompt embedding using a modified transformer block with bidirectional self-attention and cross-attention. It then upsamples the image embedding and applies an MLP-based dynamic classifier to compute the mask’s foreground probability at each location. Our approach utilizes two outputs from SAM that contains its rich segmentation knowledge. 

\section{Method}
In this section, we detail our approach for integrating SAM into the semi-supervised instance segmentation framework. First, we analyze the limitations of the pseudo-labels produced by existing teacher networks and introduce a structural distillation strategy—leveraging SAM as a meta-teacher, to enhance teacher performance in Sec.~\ref{Teacher}. Next, we describes our approach to fully integrate SAM into the student network training via pseudo-label refinement in Sec.~\ref{pseudolabel} and augmentation in Sec.~\ref{augmentation}.


\subsection{Improving teacher with structural distillation}
\label{Teacher}
\begin{figure}[t]
    \centering
    \label{fig:augmentation}
    \includegraphics[width=\linewidth]{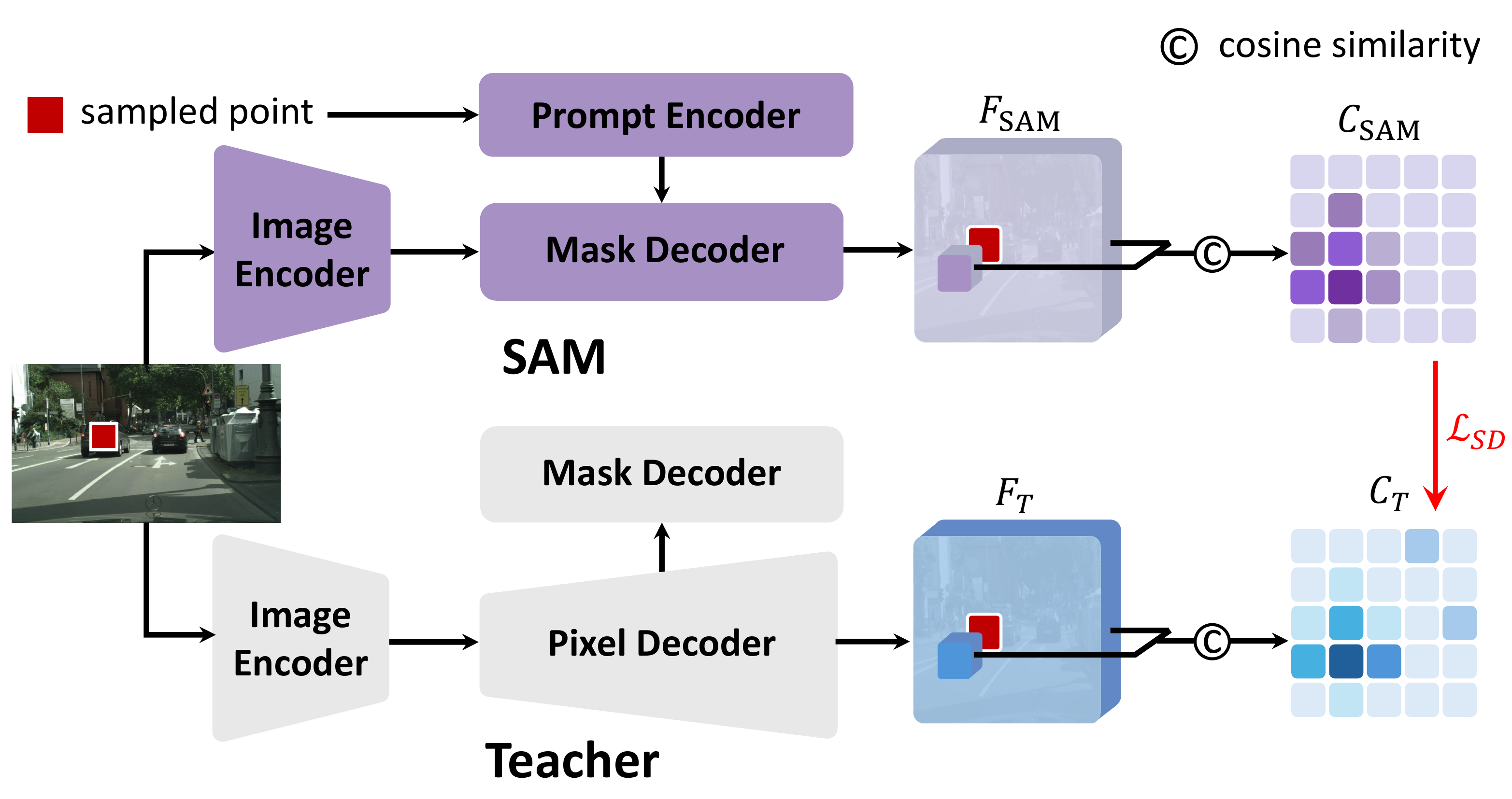}
    \caption{\textbf{Illustration of structural distillation with SAM for training the teacher.} We distill the self-similarity matrix extracted from the decoder feature of SAM to enhance the teacher for addressing under-segmentation.}
    \label{fig:sam_guided_teacher}

    \vspace{-10pt}
\end{figure}
Recent works~\cite{Berrada_2024_WACV, 10460559} have shown that a robust teacher network is essential for effective semi-supervised learning. We observe that teacher networks in existing methods tend to suffer from under-segmentation, which is mainly caused by the scarcity of labeled data and results in difficulty differentiating multiple instances, as illustrated in Fig.~\ref{fig:motivation}. Inspired by the strong localization capabilities of SAM, we propose a knowledge distillation strategy in which SAM functions as a meta-teacher, guiding the teacher network $\mathcal{F}_T$ toward finer localization. Since the teacher network is initially trained on limited labeled data, the rich representations from SAM are particularly valuable in this label-scarce regime.

A major challenge in this approach is to avoid inheriting undesirable properties from the meta-teacher, such as over-segmentation. Although SAM effectively captures fine-grained regions, its limited semantic understanding stemming from training with geometric prompts rather than semantic labels~\cite{wang2024sam} can lead to over-segmentation and suboptimal classification performance~\cite{wang2024sam, vs2024possam}. Since instance segmentation requires both accurate segmentation and robust classification, we refrain from directly minimizing the feature distance~\cite{romero2014fitnets} between SAM and the teacher. 

Instead, we design a distillation loss that focuses on imitating the structural layout of the image. 
We first extract the feature map from both SAM and teacher model. The feature map extracted from the teacher model is interpolated to match the spatial dimensions of SAM, producing $F_\mathrm{SAM}, F_T\in\mathbb{R}^{d\times H'\times W'}$.
Consequently, we compute the cosine similarity within each features, yielding self-similarity matrices $C_\mathrm{SAM}, C_T\in \mathbb{R}^{H'W' \times H'W'}$. We can interpret the slices of these similarity matrices as binary masks exhibiting the structure of the image~\cite{hong2022cost, cho2021cats,hong2022neural,hong2024unifying,hong2024unifying2,cho2022cats++,hong2021deep, cho2024cat} given query points in the image. Formally, this is defined as:
\begin{equation}
    C_\mathrm{SAM}
\;=\;
\frac{
F_\mathrm{SAM} \,\cdot\, F_{\mathrm{SAM}}
}{
\bigl\lVert F_{\mathrm{SAM}}\bigr\rVert \,\bigl\lVert F_{\mathrm{SAM}}\bigr\rVert
},
\quad
    C_T
\;=\;
\frac{
F_T \,\cdot\, F_T
}{
\bigl\lVert F_T\bigr\rVert \,\bigl\lVert F_T\bigr\rVert
}.
\end{equation}
We define structural distillation (\ourkd) loss as:
\begin{equation}
\mathcal{L}_{\mathrm{SD}}
=
\frac{1}{H'W'}\sum_{i}\rho(C_\mathrm{SAM}(i) -C_T(i)),
\end{equation}
where $\rho$ is the Huber function~\cite{huber1992robust}, and $i \in \{1, ..., H'W'\}$ is the index along the first dimension representing the query, updating the objective for the teacher as $\mathcal{L}_T = \mathcal{L}_\mathrm{lb} + \mathcal{L}_{\mathrm{SD}}$. 

One other important aspect is identifying where to distill from, in order to assure that $C_\mathrm{SAM}$ we are learning from well-captures the localized structure of the image, while avoiding over- and under-segmentation. In this regard, we further propose to distill the self-similarity matrix $C_\mathrm{SAM}$ obtained from the decoder features of SAM, instead of encoder features~\cite{wang2024sam}. Consequently, we randomly sample $P$ points, yielding $i \in \{1, ..., P\}$ and prompt the decoder with the sampled point to obtain more localized features, as illustrated in Fig.~\ref{fig:sam_guided_teacher}. 


\subsection{Refining pseudo-labels with SAM} 
\label{pseudolabel}
In addition to improving the teacher network, we propose to further boost the semi-supervised instance segmentation framework by refining the pseudo-label with SAM for minimizing the remaining error to mitigate error propagation stemming from noisy labels and to maximize the potential of the unlabeled data. Given a pseudo-label generated from the teacher, we can obtain geometric prompts to obtain refined labels from SAM. However, we find that na\"ive methods, such as selecting the center point of the mask~\cite{kirillov2023segment} or obtaining the bounding box is prone to error and often results in over-segmentation or degenerate masks. To prevent this, we introduce a simple trick by stochastically sampling multiple points as prompts to SAM.

\begin{figure}[t]
    \centering
    \includegraphics[width=\columnwidth]{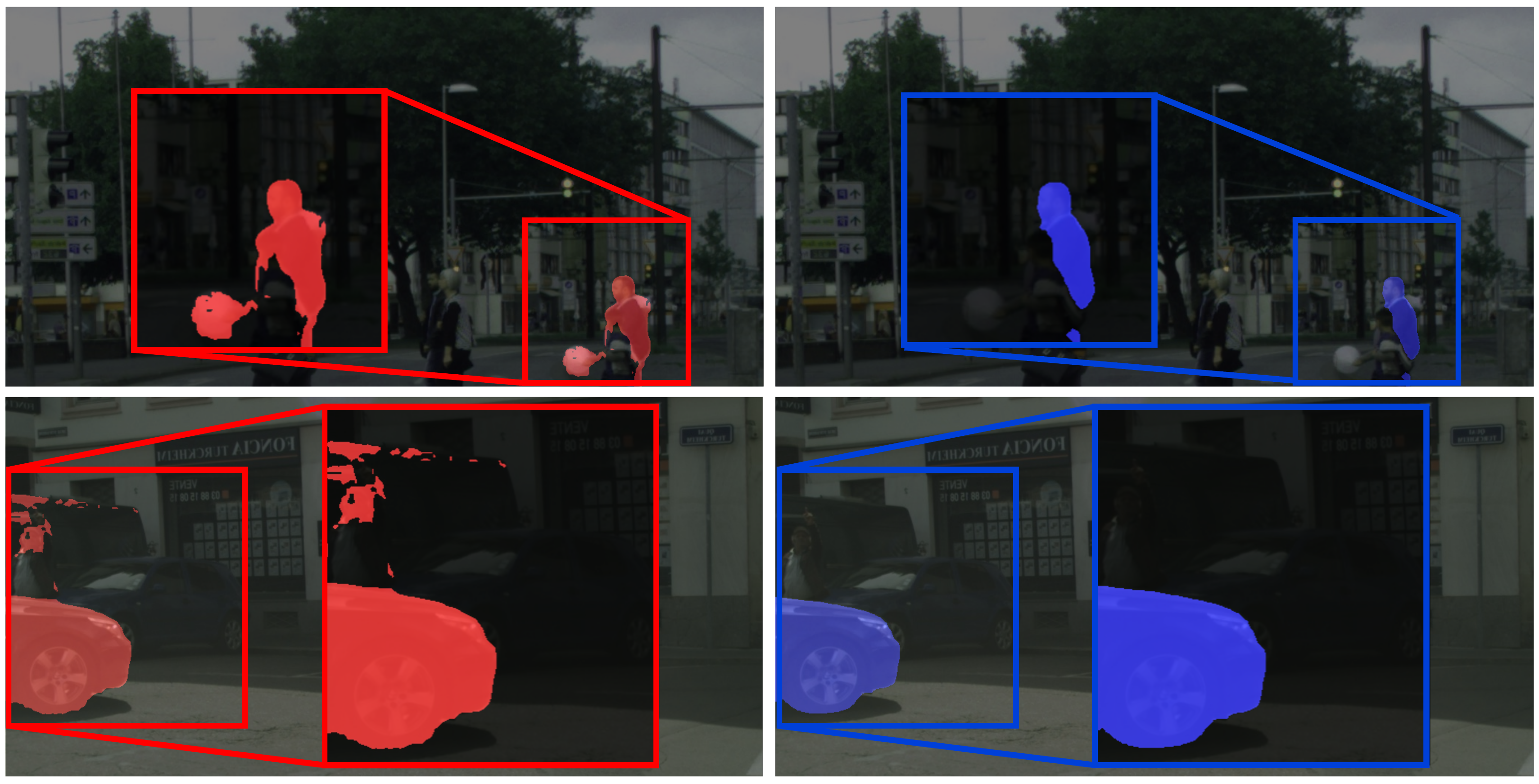}
    \caption{\textbf{Visualization of pseudo-labels before and after refinement.} We visualize pseudo-labels from the teacher network before ({\color{red}left}) and after ({\color{blue}right}) refinement. With SAM, we can refine pseudo-labels with under-segmentation, often containing noisy parts of nearby instances, into high-quality pseudo-labels.}
    \label{fig:pseudolabel}
    \vspace{-10pt}
\end{figure}

Given a pseudo-label mask $\hat{m}_u^k \in \{0, 1\}^{H \times W}$, we can also access the per-pixel probability of the mask $\tilde{\mathbf{m}}_u^k \in [0, 1]^{H \times W}$ before applying threshold to obtain binary masks. We then can obtain a probability distribution by normalizing the following:
\begin{equation}
    p(a, b) = \begin{cases}
\tilde{\mathbf{m}}^k_u, & \text{if } \hat{m}^k_u(a,b) = 1, \\
0, & \text{if } \hat{m}^k_u = 0,
\end{cases}
\end{equation}
where $a, b$ represents the spatial locations. Consequently, we obtain refined pseudo-labels by prompting SAM with $K$ points sampled from the distribution $\tilde{p}(a,b) = \frac{p(a,b)}{\sum{p(a,b)}}$. As shown in Fig.~\ref{fig:pseudolabel}, we can observe that SAM can effectively refine noisy pseudo-labels as high quality pseudo-labels. We also note that while the refined pseudo-label sometimes may not improve over the original pseudo-label due to stochasticity, we find that the student network to largely benefit from the improved samples and show consistent gains.

\subsection{Augmenting images with refined pseudo-labels} 
\label{augmentation}
The motivation for leveraging weak-to-strong consistency~\cite{sohn2020fixmatch}  in semi-supervised learning is to enforce consistent predictions under challenging conditions using strong augmentations~\cite{yun2019cutmix, cubuk2020randaugment, zhang2017mixup, devries2017improved}. This method has significantly improved tasks like semantic segmentation by utilizing techniques such as \cite{yun2019cutmix, olsson2021classmix}, which boost robustness and generalizability. However, compared to semi-supervised semantic segmentation, instance segmentation has been less explored, often relying solely on photometric augmentations. Here, we introduce Augmentation with Refined Pseudo-label (\ouraug), an augmentation strategy inspired by prior work~\cite{ghiasi2021simple}, addressing unreliable pseudo-labels through refinement to more effectively enhance performance.
\begin{figure*}[!ht]
    \centering  
    \includegraphics[width=\linewidth]{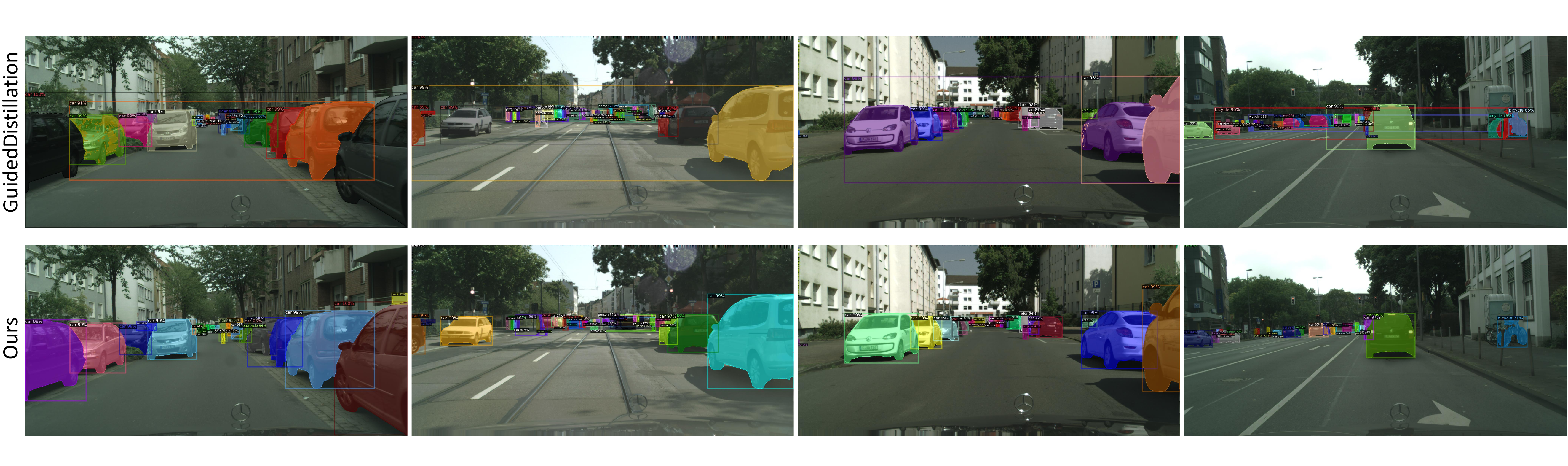}
    \vspace{-15pt}
    \caption{\textbf{Qualitative comparison on the Cityscapes dataset~\cite{cordts2016cityscapes}} using 10\% labeled data, comparing the baseline semi-supervised method GuidedDistillation~\cite{Wang_2022_CVPR} (\textbf{top}), and our approach (\textbf{bottom}). Compared to supervised training and the baseline method, our approach not only detects and segments instances more accurately but also exhibits higher discriminability between instances of the same class.}    \label{fig:main_qual_cityscapes}\vspace{-10pt}
\end{figure*}
As illustrated in Figure~\ref{fig:main_figure}, \ouraug generates synthetic images on the fly by leveraging pseudo-labels from a teacher network. Let $x_A, x_B \in D_U$ denote a randomly sampled pair of weakly augmented images from a training batch. Their refined pseudo-masks, $\{\hat{m}_A^k\}_{k=1}^{N_A}$ and $\{\hat{m}_B^k\}_{k=1}^{N_B}$, are aggregated into binary masks $M_A$ and $M_B$, where $N_A$ and $N_B$ denote the number of pseudo-label instances for $x_A$ and $x_B$, respectively. These masks are then used to bidirectionally paste detected instances between $x_A$ and $x_B$, generating synthetic images $x_{AB}, x_{BA}$ as follows:
\begin{equation}
\begin{aligned}
    x_{AB} &\leftarrow M_B \odot x_B + (1-M_B) \odot x_A, \\
    x_{BA} &\leftarrow M_A \odot x_A + (1-M_A) \odot x_B.
\end{aligned}
\end{equation}
Here, $\odot$ denotes the element-wise product between the binary mask and the image. The corresponding pseudo-labels $\hat{z}_A$ and $\hat{z}_B$ are also augmented accordingly into $\hat{z}_{AB}$ and $\hat{z}_{BA}$.  Subsequently, the student network $\mathcal{F}_S$ is trained on the photometric augmented $x_{AB}$ and $x_{BA}$.
By placing instances from paired images into each other’s contexts and backgrounds, the method introduces diverse spatial and contextual variations, including novel contexts and potential occlusions. These transformations encourage the model to perform consistently training under challenging conditions, thereby enhancing its robustness and generalization capabilities.

\section{Experiments}
\subsection{Experimental setup}
\paragraph{Datasets and evaluation metric.}
\begin{table}[t]
\centering
\begin{tabular}{l|cccc}

\toprule
\multirow{2}{*}{\textbf{Methods}} & \multicolumn{4}{c}{\textbf{Cityscapes}} \\
 & \textbf{5\%} & \textbf{10\%} & \textbf{20\%} &\textbf{30\%} \\
\midrule

DataDistillation~\cite{radosavovic2018data} & 13.7 & 19.2 & 24.6 & 27.4 \\
NoisyBoundaries~\cite{wang2022noisy} & 17.1 & 22.1 & 29.0 & 32.4 \\
PAIS~\cite{hu2023pseudo} & 18.0 & 22.9 & 29.2 & 32.8 \\
Guided Distillation~\cite{Berrada_2024_WACV} & 23.0 & 30.8 & 33.1 & 35.6 \\
Depth-Guided~\cite{chen2024depth} & \underline{23.2} & \underline{30.9} & \underline{34.1} & \underline{36.7} \\
\rowcolor{lightgray}\ours (Ours) & \textbf{30.1} & \textbf{33.3} & \textbf{37.3} & \textbf{37.8} \\
\bottomrule
\end{tabular}
\caption{\textbf{Quantitative comparison on Cityscapes.} We provide comparison of Average Precision (AP) on Cityscapes under different label ratios  with state-of-the-art methods. Results for DataDistillation is obtained from \cite{wang2022noisy}.}
\label{table:main_quan}
\end{table}

\begin{table}[t]
\centering
\begin{tabular}{l|cccc}
\toprule
\multirow{2}{*}{\textbf{Methods}} & \multicolumn{4}{c}{\textbf{COCO}} \\

 & \textbf{1\%} & \textbf{2\%} & \textbf{5\%} & \textbf{10\%}\\
\midrule
DataDistillation~\cite{radosavovic2018data} & \phantom{0}3.8 & 11.8 & 20.4 & 24.2 \\
NoisyBoundaries~\cite{wang2022noisy} & \phantom{0}7.7 & 16.3 & 24.9 & 29.2 \\
PAIS~\cite{hu2023pseudo} & 21.1 & - & 29.3 & 31.0 \\
Guided Distillation~\cite{Berrada_2024_WACV} & 21.5 & 25.3 & 29.9 & \underline{35.0} \\
Depth-Guided~\cite{chen2024depth} & \underline{22.3} & \underline{26.3} & \underline{31.5} & \textbf{35.1} \\
\rowcolor{lightgray}\ours (Ours) & \textbf{24.2} & \textbf{28.1} & \textbf{32.1} & 34.6 \\
\bottomrule
\end{tabular}
\caption{\textbf{Quantitative comparison on COCO .} We provide comparison of Average Precision (AP) on COCO under different label ratios with state-of-the-art methods. Results for DataDistillation is obtained from \cite{wang2022noisy}. 
}
\label{table:main_quan_coco}
\end{table}

We conducted our experiments on two datasets. The \textbf{Cityscapes} dataset~\cite{cordts2016cityscapes} contains 1024 x 2048 resolution driving scene images, comprising 2975 training images and 500 validation images, with pixel-level annotations for 8 semantic instance categories. For the semi-supervised setup, subsets comprising 5\%, 10\%, and 20\% of the training images were sampled and used in our experiments. The \textbf{COCO }dataset~\cite{lin2015microsoft} is a large-scale benchmark containing 118,287 images with instance segmentation annotations and is widely used in the field. For our experiments, we train on 1\%, 2\%, and 5\% subsets of the training data and evaluate the performance on the 5000-image validation set. Following previous works~\cite{he2017mask,wang2022noisy,Berrada_2024_WACV,chen2024depth}, our results are evaluated using the mask-AP metric.
\begin{figure*}[!t]
    \centering
\includegraphics[width=\linewidth]{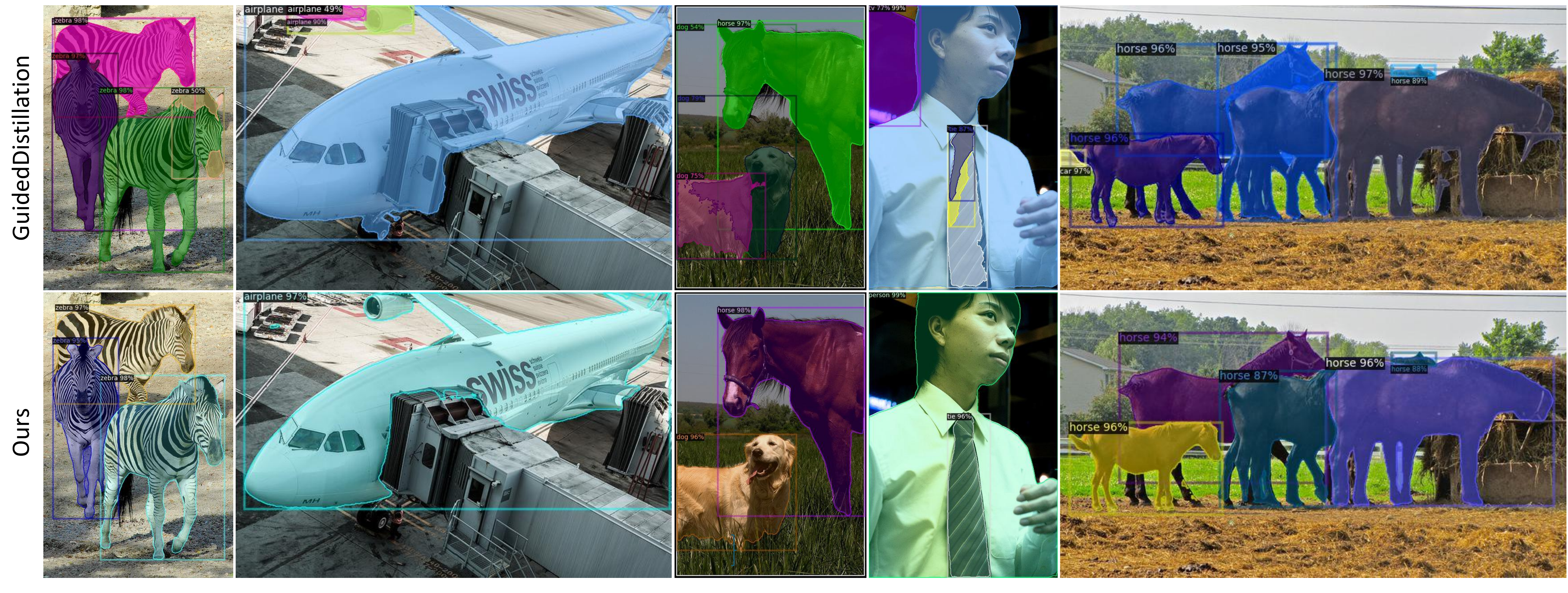}
    \caption{\textbf{Qualitative comparison on the COCO dataset~\cite{lin2015microsoft}} using 2\% labeled data, comparing the baseline semi-supervised method GuidedDistillation~\cite{Wang_2022_CVPR} (\textbf{top}), and our approach (\textbf{bottom}).}
    \label{fig:main_qual_coco}
    \vspace{-10pt}
\end{figure*}

\paragrapht{Implementation details.}
We implemented our approach based on the GuidedDistillation~\cite{Berrada_2024_WACV} codebase, using Mask2Former~\cite{cheng2022masked} with a ResNet-50~\cite{he2016deep} backbone as the instance segmentation model. For the Cityscapes dataset, we trained the model with a batch size of 16 for 90K iterations on an RTX 3090 GPU with 24GB of RAM, while for the COCO dataset, the model was trained with a batch size of 12 for 368K iterations on an RTX A6000 GPU with 48GB of RAM. The models were optimized using the AdamW optimizer with a learning rate of $10^{-4}$, a weight decay of 0.05, and a multiplier of 0.1 applied to backbone updates. The thresholds for pseudo labels were set to 0.7 for class confidence and 5 for instance size. Additionally, the EMA weight decay rate $\alpha$ was set to 0.9996, and the unsupervised loss weight $\lambda_u$ was set to 2. For pseudo-label refinement, we utilized Segment Anything Model2~\cite{ravi2024sam2} with the Hiera-L~\cite{ryali2023hiera} backbone. In our main experiments, this model was used without any additional fine-tuning on the datasets. The code will be made available upon acceptance.

\subsection{Main results}

\paragraph{Quantitative results.} Table~\ref{table:main_quan} presents the quantitative comparison results of our method with several existing works. We report the performance of models trained on the labeled data splits for the two datasets, Cityscapes and COCO, using their respective validation sets. The results demonstrate that our model outperforms previous methods, achieving state-of-the-art performance. On the \textbf{Cityscapes} dataset, our method achieves performance improvements of 6.9, 2.4, 3.2 and 1.1 points AP over the previous state-of-the-art methods for 5\%, 10\%, 20\% and 30\% labeled subsets, respectively. Compared to the teacher network, our method achieves improvements of 12.9, 10.8, 7.7, and 5.8 points AP for the same partitions. These results demonstrate the effectiveness of our proposed methods in achieving substantial performance gains even with a limited amount of labeled data. Notably, the results under the most challenging setting, 5\%, are particularly remarkable. On the \textbf{COCO} dataset, \ours achieves state-of-the-art performance across the 1\%, 2\%, and 5\% splits. In particular, it achieves improvements of 1.9 and 1.8 points AP over the previous best methods for the 1\% and 2\% splits. We observe a slight drop in the 10\% split, where we find that the stochasticity in the pseudo-label refinement could cause the drop given that the pseudo-labels are already in high-quality from having more labeled data. While could potentially address by adjusting $K$ or calibrating $\tilde{p}$ in the refinement, we highlight the substantial gains in lower label ratio splits, which aligning with the principle of semi-supervised learning for enhancing the framework with only small amount of labeled data.

\paragrapht{Qualitative results.} We provide qualitative examples for both Cityscapes and COCO in Figure~\ref{fig:main_qual_cityscapes} and Figure~\ref{fig:main_qual_coco}. When comparing the results of our method to those from comparative baseline, GuidedDistillation, we observe that our model is trained to align more closely with the goals of instance segmentation. In the case of GuidedDistillation, both datasets exhibit instances where multiple objects of the same semantic class are not properly separated, leading to the inclusion of multiple instances in a single mask proposal. In contrast, our model not only demonstrates higher performance in accurately distinguishing between instances but also achieves improved segmentation accuracy. We contribute our gains particularly to the careful adoption of SAM, effectively addressing under- and over-segmentation seen in compared baseline. We provide further qualitative results in the supplementary materials.

\subsection{Ablation studies}
We provide ablation studies for validating our approach and our design choices. All experiments were performed with the Mask2Former model with ResNet-50 backbone, consistent with the setup used in the main experiments. We report the results of the best model obtained over 45K training iterations from Cityscapes dataset using the 10\% partition of the labeled data if not specificed.



\begin{table}[t]
\centering
\resizebox{\columnwidth}{!}{%
\begin{tabular}{c|cccc|cccc}
\toprule
\multirow{2}{*}{\textbf{SD}} & \multicolumn{4}{c|}{\textbf{Cityscapes}} & \multicolumn{4}{c}{\textbf{COCO}} \\
 & \textbf{5\%} & \textbf{10\%} & \textbf{20\%} & \textbf{30\%} & \textbf{1\%} & \textbf{2\%} & \textbf{5\%} & \textbf{10\%} \\
\midrule
\ding{55} & 14.9 & 19.0 & 27.6 & 29.9 & 13.5 & 19.5 & 25.8 & 30.1\\
\rowcolor{lightgray} \ding{51} & \textbf{17.2} & \textbf{22.5} & \textbf{29.6} & \textbf{32.0} & \textbf{15.1} & \textbf{20.3} & \textbf{26.1} & \textbf{30.3} \\
\bottomrule
\end{tabular}
}




\caption{\textbf{Effects of the structural distillation (\ourkd) loss for pre-training the teacher.} We compare our improved teacher with the structural distillation loss to the baseline teacher reproduced from \cite{Berrada_2024_WACV}, which is trained only with $\mathcal{L}_\mathrm{lb}$.}
\label{table:teacher_quan}
\vspace{-10pt}
\end{table}
\begin{table}[h]
\centering
\begin{minipage}[t]{0.455\columnwidth}
\centering
\resizebox{\linewidth}{!}{%
    \begin{tabular}{cc|c}
    \toprule
    \multicolumn{2}{c|}{$\mathcal{L}_{SD}$} & \multirow{2}{*}{AP} \\
    teacher & student & \\
    \midrule
    \grayx & \grayx & 30.1 \\
    \rowcolor{lightgray} \blackcheck & \textcolor{darkgray}{\ding{55}} &\textbf{32.8} \\
    \grayx & \blackcheck & 30.7 \\
    \blackcheck & \blackcheck & 29.4  \\ 
    \bottomrule
    \end{tabular}
}
\caption{\textbf{Ablation on $\mathcal{L}_{SD}$ in different training stages.}}
\label{tab:sd1}
\end{minipage}\hfill
\begin{minipage}[t]{0.505\columnwidth}
\centering
\resizebox{\linewidth}{!}{%
\begin{tabular}{cc|c}
    \toprule
    \multirow{2}{*}{$F_{\mathrm{SAM}}$} & distill. & \multirow{2}{*}{AP} \\
    & loss & \\
    \midrule
    - & - & 19.0 \\
    encoder & feature & 21.1 \\
    encoder & structural & 21.4 \\  
    \rowcolor{lightgray} decoder & structural &  \textbf{22.4} \\
    \bottomrule
\end{tabular}
}
\caption{\textbf{Ablation on $F_\mathrm{SAM}$ and distillation loss.}}
\label{tab:sd2}
\end{minipage}

\vspace{-10pt}
\end{table}

\paragrapht{Effects of the \ourkd loss for the teacher.}
Tab~\ref{table:teacher_quan} presents the performance of the teacher network enhanced with the proposed \ourkd loss across varying label ratios on the Cityscapes and COCO datasets. The baseline refers to a Mask2Former~\cite{cheng2022masked} trained with $\mathcal{L}_{lb}$ for different label ratios, following GuidedDistillation~\cite{Berrada_2024_WACV}. On Cityscapes with 5\% labeled data, the SD loss yields an improvement of +2.3 AP over the baseline, with consistent gains of +3.5 AP, +2.0 AP, and +2.1 AP observed at 10\%, 20\%, and 30\% label ratios, respectively. Similarly, on COCO, the SD loss achieves a notable gain of +1.6 AP under the 1\% label setting, with improvements maintained across higher label ratios. These results indicate that incorporating the \ourkd loss effectively enhances the pre-training process across all splits with especially higher gains for lower label ratio splits, demonstrating its efficacy in label-scarce settings.

\begin{figure}[t!]
    \centering
    \includegraphics[width=\linewidth]{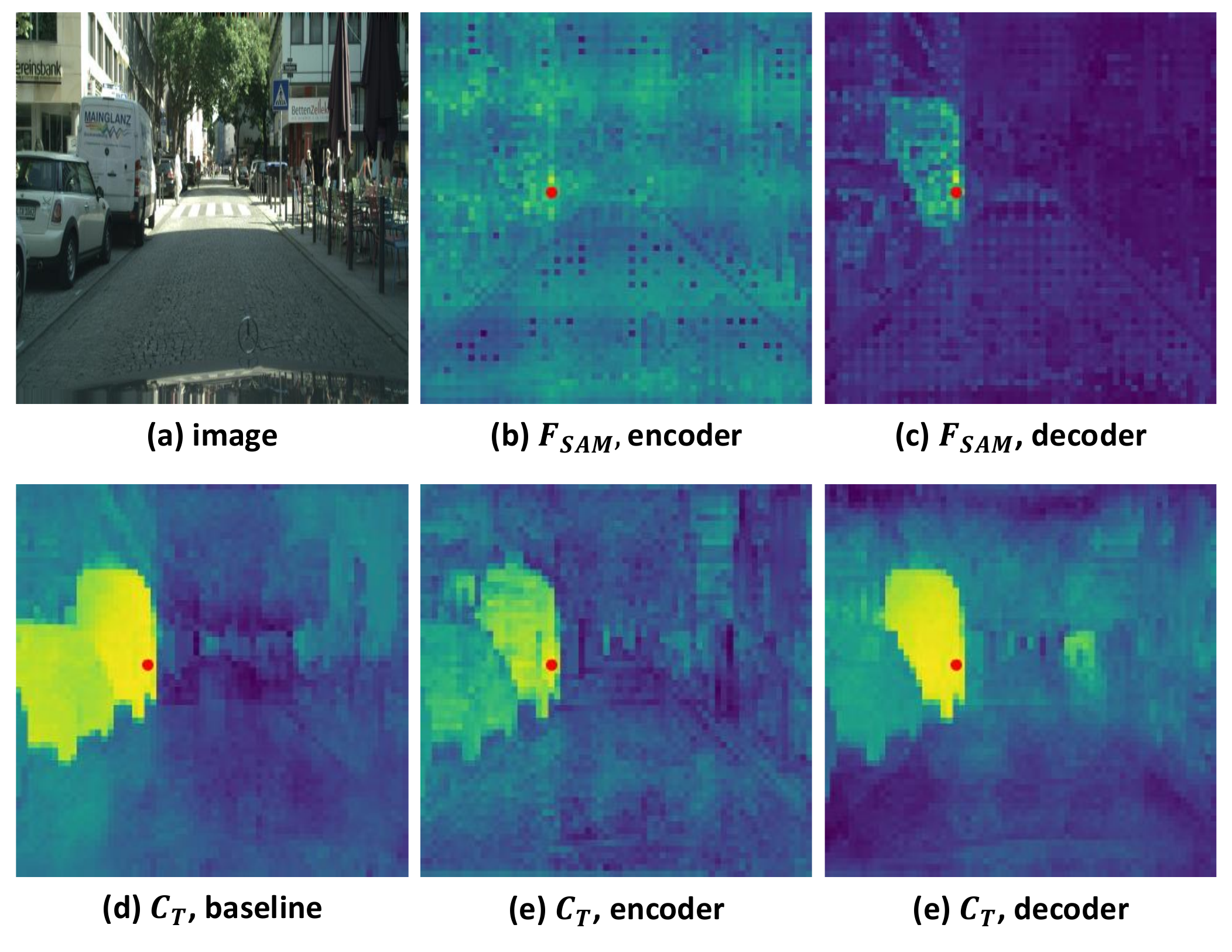}
    \caption{\textbf{Visualization of self-similarity matrices $C$.} Given an image (a), we visualize $C_\mathrm{SAM}$ with encoder (b) and decoder (c) features for $F_\mathrm{SAM}$. We also visualize $C_T$ for baseline teacher(d) , as well as the corresponding teachers trained with $\mathcal{L}_{SD}$ with $C_\mathrm{SAM}$ from encoder (e) and decoder (f).}
    \label{fig:featuremap}
    \vspace{-15pt}
\end{figure}

\paragrapht{Ablation of \ourkd loss in different training stages.}
Table~\ref{tab:sd1} presents an ablation study on the application of the \ourkd loss, specifically examining the effect of employing the structural distillation loss, $\mathcal{L}_\mathrm{SD}$, at different training stages. When $\mathcal{L}_\mathrm{SD}$ is applied exclusively during teacher training or solely during student training, both configurations yield improvements. However, the teacher-only configuration delivers a substantially higher overall gain, underscoring the critical importance of a robust teacher network. Surprisingly, applying $\mathcal{L}_\mathrm{SD}$ to both the teacher and student stages results in a performance drop after the burn-in iterations~\cite{Berrada_2024_WACV}, suggesting that the burn-in stage to be problematic.

\paragrapht{Ablation on design choices for \ourkd loss.}
In Table~\ref{tab:sd2}, we present ablations investigating the design choices of our \ourkd loss by comparing different strategies for applying the structural distillation loss $\mathcal{L}_\mathrm{SD}$ in the teacher network. When using feature distillation~\cite{romero2014fitnets}, we minimize the Euclidean distance between $F_\mathrm{SAM}$ and $F_T$. Although all configurations yield noticeable improvements compared to the baseline without it, we observe that employing structural distillation leads to better teacher performance than feature distillation. Moreover, utilizing decoder-derived features for $F_{\mathrm{SAM}}$ results in an additional gain, verifying our choices.

\begin{table}[t!]
\centering
\begin{tabular}{cccc|cc}
\toprule
& \ourkd & PR & \ouraug & AP\\ 
\midrule
\textbf{(I)} & \grayx& \grayx& \grayx& 28.4\\ 
\textbf{(II)}&\ding{51}& \grayx& \grayx& 29.2 \\
\textbf{(III)}&\grayx& \grayx& \blackcheck& 27.8 \\ 
\textbf{(IV)}&\blackcheck& \blackcheck& \grayx& 32.3\\ 
\textbf{(V)}&\blackcheck& \grayx& \blackcheck & 31.0 \\ 
\rowcolor{lightgray} \textbf{(VI)} & \blackcheck& \blackcheck& \blackcheck& \textbf{32.8} \\ 
\bottomrule
\end{tabular}
\caption{\textbf{Component analysis.} We conduct ablation study on our key components structural distillation (SD), pseudo-label refinement (PR), and augmentation (\ouraug).}
\label{table:ablation_component}
\vspace{-15pt}
\end{table}
Figure~\ref{fig:featuremap} provides visualizations of $C_\mathrm{SAM}$ and $C_T$ in the context of the structural distillation loss. As shown in (b-c), the $C_\mathrm{SAM}$ obtained from the decoder demonstrates improved localization and maintains high similarity within the target instance at an appropriate level of granularity. Additionally, panels (e-f) illustrate the corresponding teacher self-similarity matrix $C_T$ from a teacher trained with $\mathcal{L}_\mathrm{SD}$, which, when compared to the baseline in panel (d) (i.e., without $\mathcal{L}_\mathrm{SD}$), clearly distinguishes the similarity across different instances within the same class.
\vspace{-2pt}
\paragrapht{Component analysis.} 
We conduct ablation experiments to evaluate the effectiveness of the main components of our method: structural distillation loss (SD), pseudo-label refinement (PR) and the proposed augmentation strategy, \ouraug. As shown in Table~\ref{table:ablation_component}, the performance improvement of the teacher model trained with $\mathcal{L}_\mathrm{SD}$ loss positively contributes to the learning of the student network (\textbf{II}). Furthermore, applying pseudo-label refinement leads to a significant performance boost (\textbf{IV}), demonstrating the effectiveness of incorporating SAM into a semi-supervised instance segmentation framework. While \ouraug alone can negatively affect student training (\textbf{III})—likely due to noise from unrefined pseudo-labels—its combination with a teacher trained under SD still yields performance gains (\textbf{V}), suggesting improved pseudo-label generation. Moreover, combining \ouraug with pseudo-label refinement achieves the best performance (\textbf{VI}), indicating that these two methods work synergistically to enhance learning.

\section{Conclusion}
In this work, we propose \ours, a novel semi-supervised instance segmentation framework that integrates the SAM through structured distillation, pseudo-label refinement, and data augmentation. By selectively leveraging precise localization capability of SAM while mitigating its over-segmentation tendency, our approach significantly improves the teacher-student framework. Extensive experiments demonstrate state-of-the-art performance across benchmarks, highlighting the effectiveness of our method in enhancing semi-supervised instance segmentation.

{
    \small
    \bibliographystyle{ieeenat_fullname}
    \bibliography{main}
}

\clearpage
\renewcommand{\thesection}{\Alph{section}}
\renewcommand{\thefigure}{A.\arabic{figure}}
\renewcommand{\thetable}{A.\arabic{table}}

\setcounter{section}{0}
\setcounter{figure}{0}
\setcounter{table}{0}
\setcounter{page}{1}
\maketitlesupplementary

\section{More Qualitative Results}
\begin{figure*}[t]
    \centering
    \vspace{-10pt}
    \includegraphics[width=\linewidth]{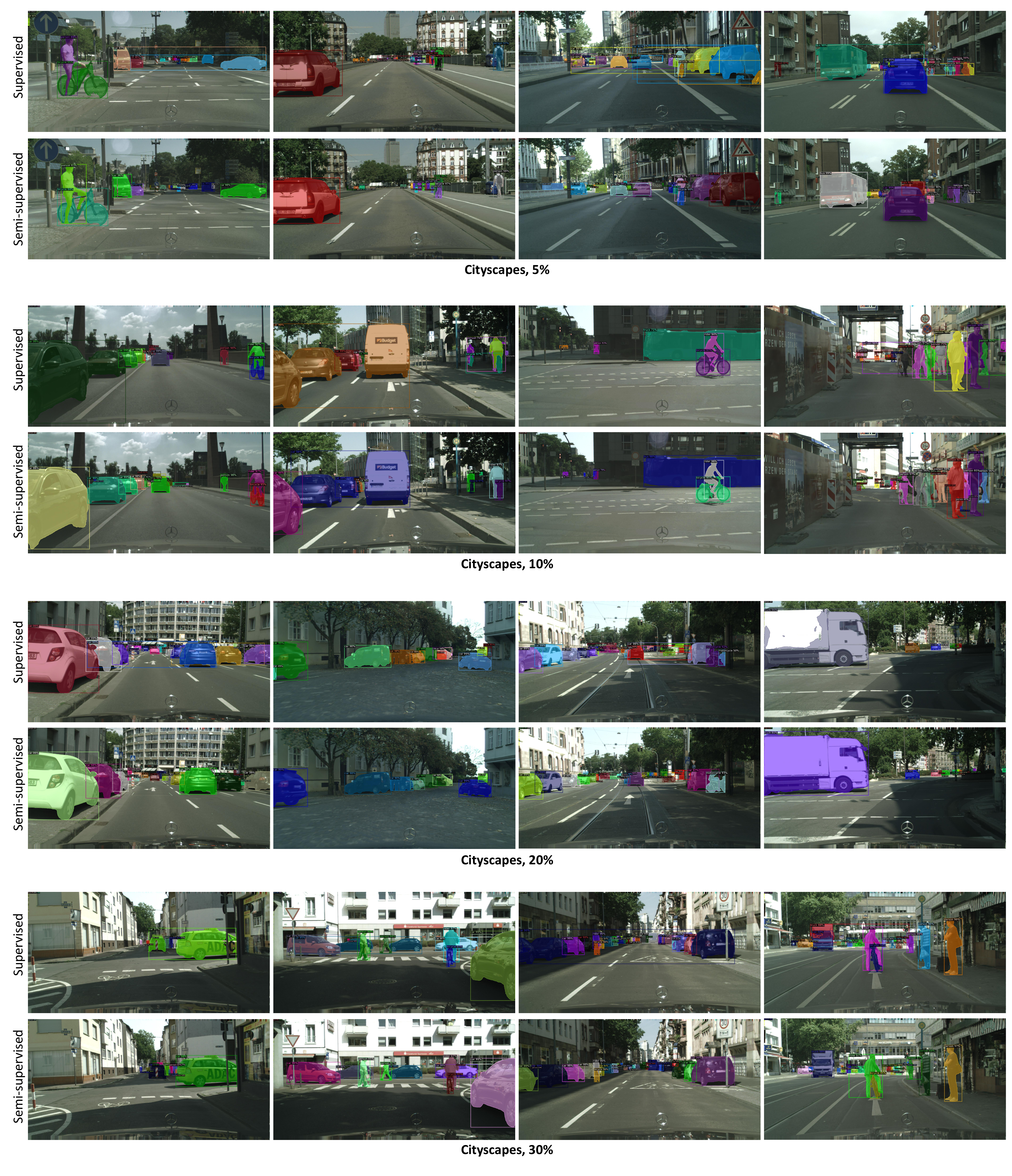}
    \vspace{-15pt}
    \caption{\textbf{Qualitative results on Cityscapes under different labeled data settings.} Predictions from supervised training (top) and our semi-supervised approach (bottom) across different labeled data settings. "Supervised” refers to the pretrained teacher network, while “semi-supervised” denotes the student model trained jointly on both labeled and unlabeled data. } 
    \label{fig:cityscapes_quals_supp}
    \vspace{-10pt}
\end{figure*}

\begin{figure*}[t]
    \centering
    \vspace{-10pt}
    \includegraphics[width=\linewidth]{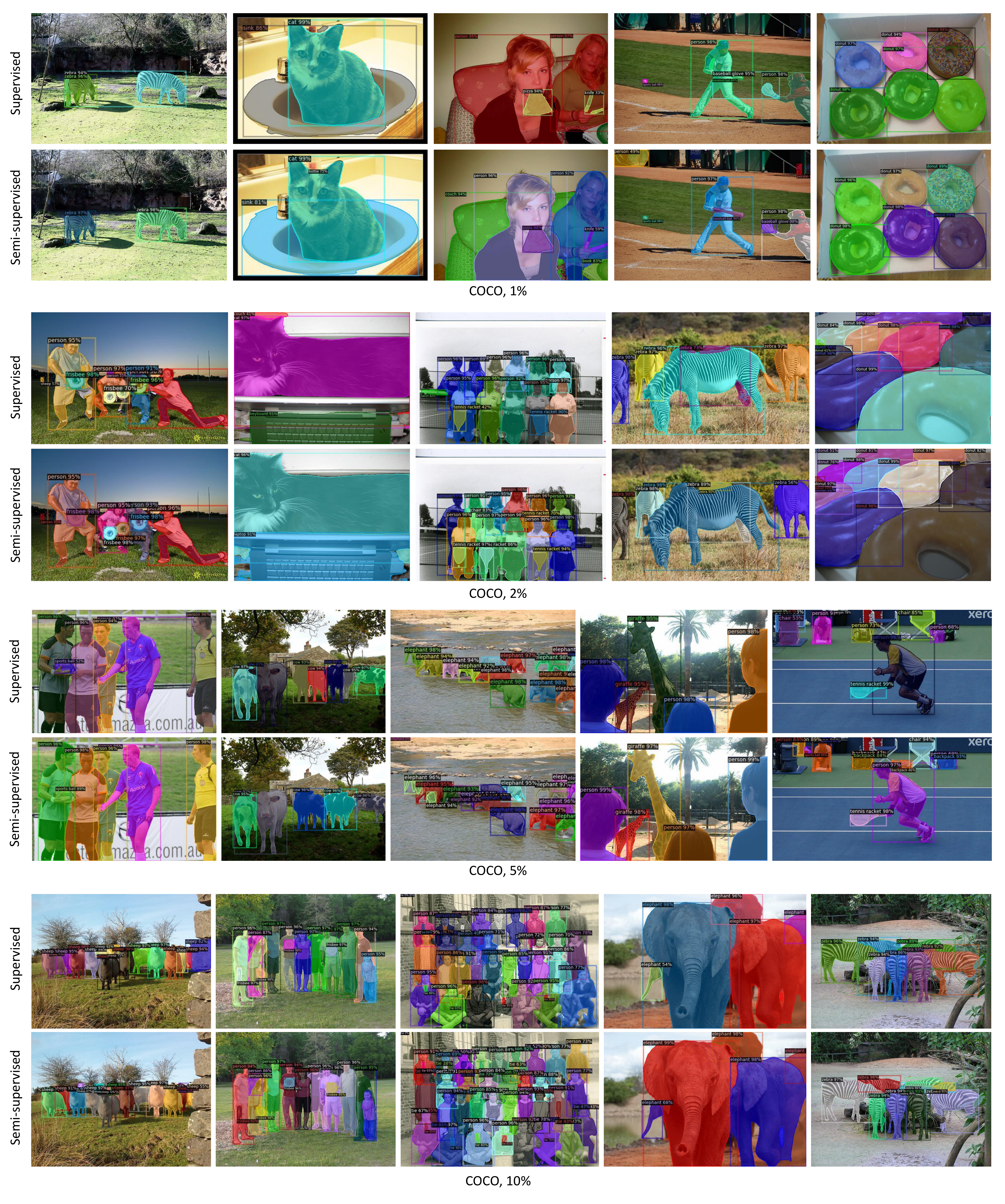}
    \vspace{-15pt}
    \caption{\textbf{Qualitative results on COCO under different labeled data settings.} Predictions from supervised training (top) and our semi-supervised approach (bottom) across different labeled data settings.} 
    \label{fig:coco_quals_supp}
    \vspace{-10pt}
\end{figure*}

\begin{figure*}[t]
    \centering
    \vspace{-10pt}
    \includegraphics[width=\linewidth]{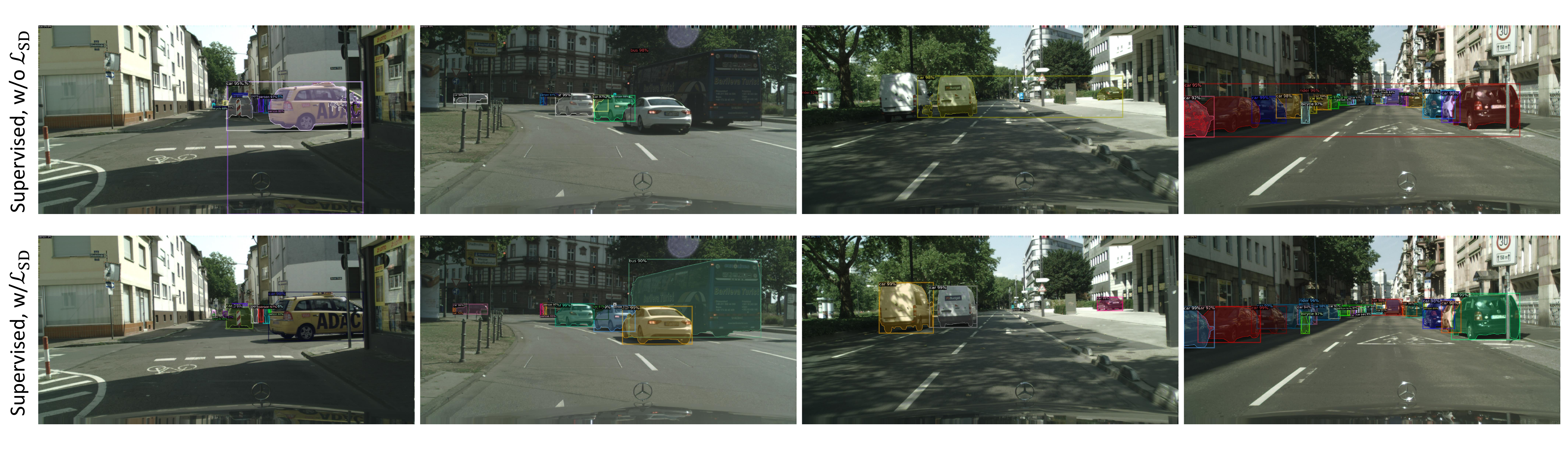}
    \vspace{-15pt}
    \caption{\textbf{Qualitative comparison between the improved teacher model, enhanced by structural distillation and trained on 20\% of the labeled data, and the baseline model on Cityscapes.}} 
    \label{fig:teacher_sd_comparison}
    \vspace{-10pt}
\end{figure*}

\paragraph{Extended qualitative results.}
We present extended qualitative results of \ours on Cityscapes~\cite{cordts2016cityscapes} in ~\ref{fig:cityscapes_quals_supp} and for COCO~\cite{lin2015microsoft} in ~\ref{fig:coco_quals_supp}. The results demonstrate that our approach consistently achieved improvements over the supervised teacher network across all experimental settings.
\paragraph{Qualitative comparison of the improved teacher with structural distillation.}
In addition to the quantitative results presented in Tab. 1 of the main paper, we provide qualitative evidence in ~\ref{fig:teacher_sd_comparison} to illustrate the improvements achieved by the teacher model trained with structural distillation (SD). \ref{fig:teacher_sd_comparison} demonstrates that the supervised model with the additional SD loss $\mathcal{L}_\mathrm{SD}$ detects objects more effectively and reduces instances where multiple instance masks are merged into a single pseudo-label compared to the baseline model without $\mathcal{L}_\mathrm{SD}$.

\section{Examples of augmented images with refined pseudo-labels}
\begin{figure*}[t]
    \centering
    \vspace{-10pt}
    \includegraphics[width=\linewidth]{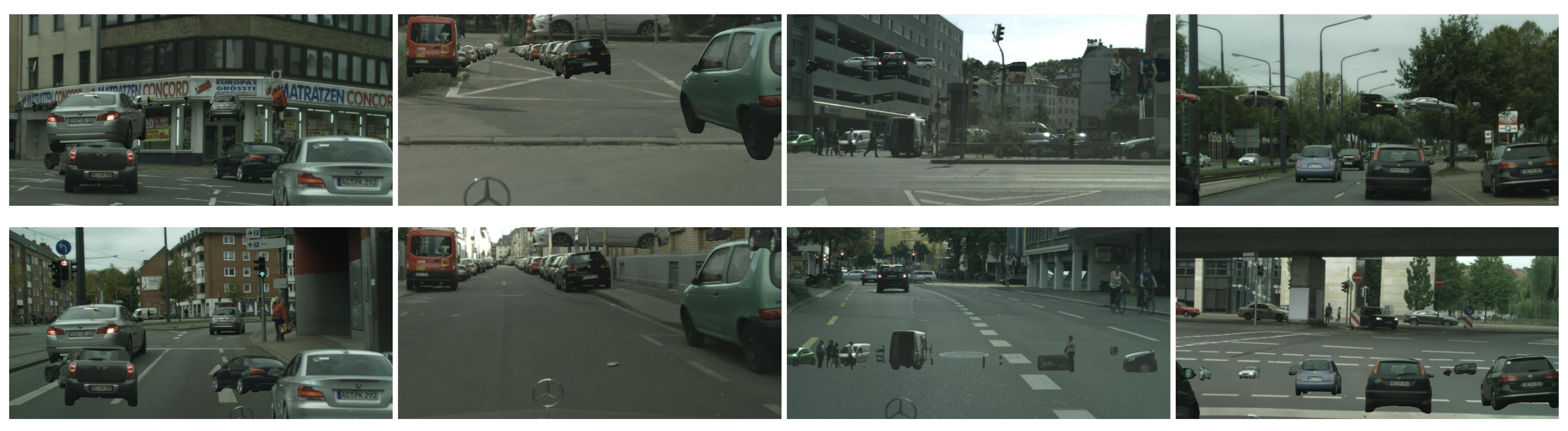}
    \vspace{-15pt}
    \caption{\textbf{Visualization of augmented samples with refined pseudo-labels}}
    \label{fig:aug_samples}
    \vspace{-10pt}
\end{figure*}

In Fig.~\ref{fig:aug_samples}, we present sample outputs of our proposed Refined Instance Mixing (\textbf{\ouraug}). Pseudo-label masks are initially generated from teacher predictions and then refined using SAM, yielding higher-quality pseudo-labels. Building upon these enhanced labels, \ouraug craft synthetic data by blending instances from paired images, thereby introducing diverse spatial and contextual variations such as novel backgrounds and potential occlusions. This augmentation strategy encourages consistent model performance under challenging conditions and fosters improved robustness and generalization to a wide range of transformations.

\section{Additional Analysis}
\paragraph{Analysis on pseudo-label quality.} 
Analysis of pseudo-label quality for the original teacher prediction, as provided in Fig. 1 of the main paper, was conducted on the Cityscapes validation set. The segmentation quality (SQ) was quantified using the mean IoU of true positive labels, where a prediction was considered a positive label if it shared the same class with the ground truth and had an IoU exceeding 0.5. Class accuracy (CA) was computed as the ratio of correctly matched predictions (true positives) to the total number of predictions, with matched pairs defined as predictions exceeding an IoU threshold of 0.5. Notably, we did not utilize the region quality metric commonly employed in panoptic quality (PQ) for evaluating class accuracy, as its computation considers false negatives, leading to the inclusion of undetected pseudo-labels and making accurate assessment difficult. Building on this analysis, we evaluated teacher predictions refined through structural distillation, confirming its effectiveness in improving pseudo-label quality metrics and addressing the identified challenges. Tab~\ref{tab:pseudo_label_comparison} indicates that structural distillation effectively enhances both metrics used to evaluate pseudo-label quality, thereby substantively addressing the challenges we discussed.
\begin{table}[htbp]
\centering
\begin{tabular}{c|cc}
\toprule
& Baseline & Baseline+\ourkd \\ 
\midrule
CA & 96.9 & \textbf{97.3} \\ 
SQ & 47.7 & \textbf{49.4} \\ 
\bottomrule
\end{tabular}
\caption{\textbf{Comparison of pseudo-label quality analysis.}}
\label{tab:pseudo_label_comparison}
\end{table}

\paragrapht{Analysis on prompt types for pseudo-label refinement.}
In Tab~\ref{table:able_prompt}, we present the results of applying different SAM prompt types (bounding box, mask, and point) to our method. The results show that multiple point prompts offer the highest performance, aligning with our proposed approach. While bounding boxes follow closely, they can introduce ambiguity when multiple objects appear within a single box. Single-point prompts can lead to degraded performance due to SAM’s over-segmentation tendencies. Furthermore, as discussed in prior work~\cite{dai2023samaug, zhang2023personalize}, relying on mask prompts may lower mask quality and thus negatively affect training.

\begin{table}[h!]
\centering
\begin{tabular}{l|c}
\toprule
Prompt Type & AP \\
\midrule
Bounding Box & 32.1 \\
Mask &  24.3\\
Single point & 30.4 \\
$K$-sampled points (Ours) & \textbf{32.8} \\
\bottomrule
\end{tabular}
\caption{\textbf{Performance comparison across different prompt type configurations.}}
\label{table:able_prompt}
\end{table}

\clearpage
\clearpage

\end{document}